\documentclass[10pt,twocolumn,letterpaper]{article}

\usepackage{3dv}
\usepackage{times}
\usepackage{epsfig}
\usepackage{graphicx}
\usepackage{amsmath}
\usepackage{amssymb}

\usepackage[pagebackref=true,breaklinks=true,letterpaper=true,colorlinks,bookmarks=false]{hyperref}

\threedvfinalcopy 

\def\threedvPaperID{118} 

\ifthreedvfinal\fi
\begin{document}

\title{Depth Fields: Extending Light Field Techniques to Time-of-Flight Imaging}

\author{Suren Jayasuriya\\
Cornell University\\
{\tt\small sj498@cornell.edu}
\and
Adithya Pediredla\\
Rice University\\
{\tt\small adithya.k.pediredla@rice.edu}
\and
Sriram Sivaramakrishnan\\
Cornell University\\
{\tt\small ss2462@cornell.edu}
\and
Alyosha Molnar\\
Cornell University\\
{\tt\small molnar@ece.cornell.edu}
\and
Ashok Veeraraghavan\\
Rice University\\
{\tt\small vashok@rice.edu}
}

\maketitle

\begin{abstract}
A variety of techniques such as light field, structured illumination, and time-of-flight (TOF) are commonly used for depth acquisition in consumer imaging, robotics and many other applications. Unfortunately, each technique suffers from its individual limitations preventing robust depth sensing. In this paper, we explore the strengths and weaknesses of combining light field and time-of-flight imaging, particularly the feasibility of an on-chip implementation as a single hybrid depth sensor. We refer to this combination as depth field imaging. Depth fields combine light field advantages such as synthetic aperture refocusing with TOF imaging advantages such as high depth resolution and coded signal processing to resolve multipath interference. We show applications including synthesizing virtual apertures for TOF imaging, improved depth mapping through partial and scattering occluders, and single frequency TOF phase unwrapping. Utilizing space, angle, and temporal coding, depth fields can improve depth sensing in the wild and generate new insights into the dimensions of light's plenoptic function. 
\end{abstract}


\section{Introduction}

The introduction of depth sensing to capture 3D information has led to its ubiquitous use in imaging and camera systems, and has been a major focus of research in computer vision and graphics. Depth values enable easier scene understanding and modeling which in turn can realize new computer vision systems and human-computer interaction. Many methods have been proposed to capture depth information such as stereo, photometric stereo, structured illumination, light field, RGB-D, and TOF imaging. 

However depth cameras typically support only one depth sensing technology at a time which limits their robustness and flexibility. Each imaging modality has its own advantages and disadvantages for attributes such as on-chip implementation, cost, depth resolution, etc that are summarized in Table~\ref{table}. We argue that hybrid 3D imaging systems which utilize two or more depth sensing techniques can overcome these individual limitations. Furthermore, a system that combines modalities with an on-chip implementation would be cost effective and mass producible, allowing ubiquitous robust depth sensing.

\begin{table}
\begin{center}
   \includegraphics[width=1\linewidth]{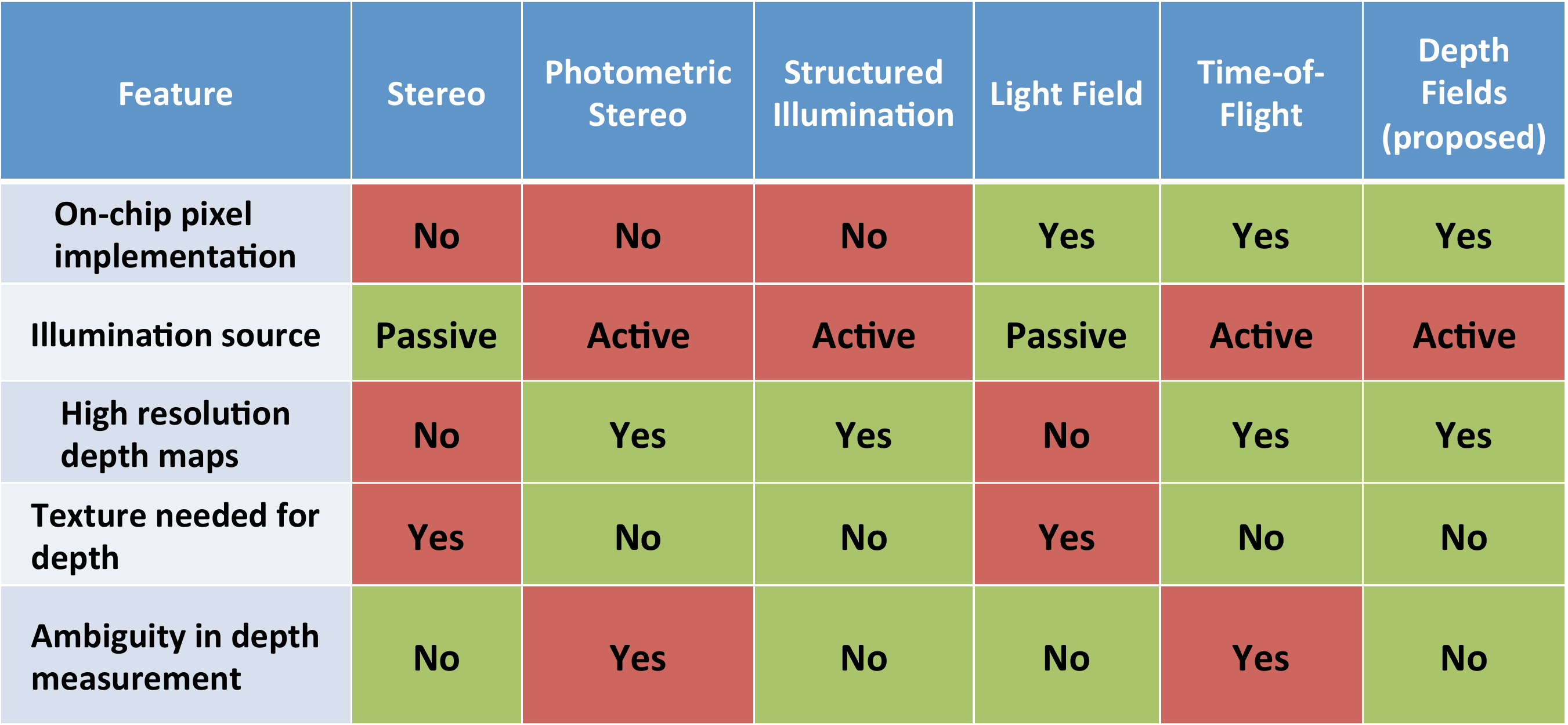}
\end{center}
\caption{Table that summarizes the relative advantages and disadvantages of different depth sensing modalities including the proposed depth fields.}
\label{table}
\end{table} 

We propose combining light field and TOF imaging into a hybrid 3D imaging system. This system inherits light field advantages such as post-capture digital refocusing with TOF advantages of high resolution depth information and the mitigated multipath interference using coded signals. Further, light field and TOF imaging both have been implemented on-chip~\cite{Bamji2004, ng2005light}, and we can design hybrid pixel structures to combine both modalities on-chip as well. Each modality has its relative disadvantages: depth from light fields require textured surfaces and is dependent on object distance for disparity, and single frequency TOF imaging suffers from phase wrapping and is limited to small aperture cameras with low shutter speeds. However, we show that combining light field and TOF imaging can alleviate all of these limitations.

We call this extension of spatio-angular information captured traditionally by light fields to TOF depth maps as \textbf{depth field imaging}. Our main contributions include:

\begin{itemize}
\item Formulation of depth field imaging as an extension of the light field framework for TOF imaging
\item Methods to capture depth fields using camera arrays and single-shot camera systems.
\end{itemize}
We show that capturing depth fields leads to many new applications that improve robust depth sensing in the wild including:
\begin{itemize}
\item Digital refocusing of depth images and extended depth of field.
\item Phase unwrapping for single frequency TOF imaging.
\item Depth imaging through partial occluders.
\item Depth imaging and refocusing past scattering media.
\end{itemize}

A larger vision for introducing depth fields is to have a layer of post-capture control for depth sensing which can combine synergistically with higher level algorithms such as structure from motion (SfM)~\cite{Thrun2000}, Simultaneous Localization and Mapping (SLAM)~\cite{Thrun2003}, and reconstructions from collections of online images~\cite{Snavely2006} used for 3D reconstruction and scene modeling/understanding. 



\section{Related Work}
We survey related work in light field, TOF, and fusion algorithms for depth imaging to show the context of depth sensing technologies that depth field imaging relates to.

\textbf{Light Field Imaging} captures 4D representations of the plenoptic function parametrized by two spatial coordinates and two angular coordinates, or equivalently as the space of non-occluded rays in a scene~\cite{gortler1996lumigraph, levoy1996light}. Light fields are used for image-based rendering and modeling, synthesizing new viewpoints from a scene, and estimating depth from epipolar geometry. In the context of cameras, light fields have been captured by using mechanical gantries~\cite{Levoy2004gantry} or large dense camera arrays~\cite{wilburn2005high}, or by single-shot methods including microlenses~\cite{adelson1992single, ng2005light}, coded apertures~\cite{Levin2007codedaperture}, transmission masks~\cite{veeraraghavan2007dappled}, or diffraction gratings~\cite{hirsch2014switchable}. Light fields can extend the depth of field and use digital refocusing to synthesize different apertures in post-processing~\cite{ng2005light}, thus enabling a level of software control after the photograph has been taken. We will exploit this control in depth field imaging.

\textbf{Time-of-Flight Imaging} works by encoding optical path length traveled by amplitude modulated light which is recovered by various devices including photogates and photonic mixer devices~\cite{Payne2014, Bamji2004, Lange2000, Schwarte1997}. While yielding high resolution depth maps, single frequency TOF suffers from limitations including phase wrapping ambiguity and multipath interference caused by translucent objects and scattering media. Proposed techniques to overcome these limitations include phase unwrapping with multifrequency methods~\cite{payne2009multiple}, global/direct illumination separation~\cite{kadambidemultiplexing, Wu2014global}, deblurring and superresolution~\cite{xiao2015defocus}, and mitigating multipath interference with post-processing algorithms~\cite{bhandari2014sparse, bhandari2014resolving}. Recently, new temporal coding patterns for these sensors help resolve multiple optical paths to enable seeing light in flight and looking through turbid media~\cite{heide2013low, heide2014imaging, kadambi2013coded}. Similar to this paper, camera systems have been proposed to fuse together TOF + stereo~\cite{zhu2008fusion}, TOF + photometric stereo~\cite{ti2015simultaneous}, and TOF + polarization~\cite{Kadambi2015iccv}.

 \textbf{Fusion of depth maps and intensity images} has been used to enable 3D reconstruction by explicit feature detection~\cite{Henry2012rgbd, Huhle2010}. Real-time interaction for camera tracking and 3D reconstruction have been demonstrated via KinectFusion~\cite{Izadi2011}. While conceptually similar to depth fields by acquiring per-pixel values of depth and intensity, these fusion methods do not systematically control the spatio-angular sampling or transcend the traditional capture tradeoffs between aperture and depth of field for depth imaging. In this way, we hope that depth field algorithms can serve as the foundation upon which fusion algorithms can improve their reconstruction quality, leading vertical integration from camera control all the way to high level scene modeling and understanding. 

\begin{figure*}
\begin{center}
\includegraphics[width=1\linewidth]{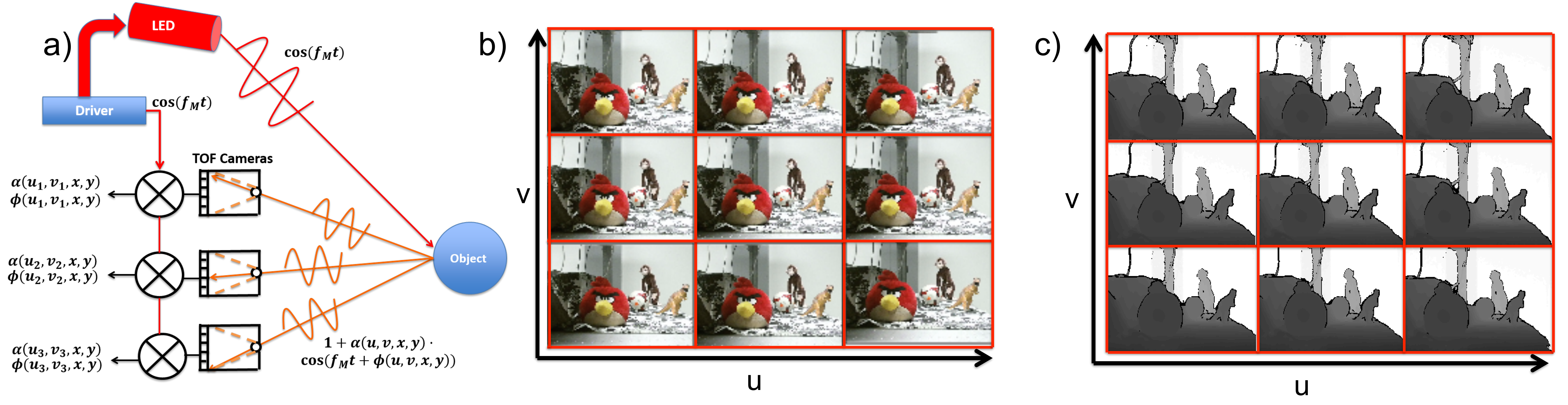} 
\end{center}
   \caption{(a) Capturing a depth field conceptually using an array of TOF cameras, (b) albedo as 4D function of (u,v,x,y), (c) phase corresponding to TOF depth values as a 4D function of (u,v,x,y)}
\label{conceptfigure}
\end{figure*}


\section{Depth Fields}
In this section, we combine the mathematical formulations of light field and TOF imaging into the concept of a depth field. We show both how to capture these fields and how to invert the forward model to recover light albedo, defined as the reflectance value of an object with respect to the active illumination, and depth as a function of 2D spatial coordinates and 2D angular coordinates. This approach is similar to Kim et al.~\cite{Kim2013lightfield} who capture depth maps for different perspective views, but they do not use TOF imaging or show applications such as digital refocusing or depth mapping through partial/scattering occluders. 

To describe the forward model for capturing depth fields, we first briefly discuss the forward models for light field and TOF imaging. 
\subsection{Light Fields}
Light fields are commonly parameterized by the two plane model $l(u,v,x,y)$ where $(u,v)$ is the angular coordinates at the lens plane, and $(x,y)$ are the spatial coordinates of the sensor plane~\cite{levoy1996light}. The output of this function represents the radiance of the ray parametrized by its intersection with the two planes. The forward model for light field capture has been modeled in~\cite{Wetzstein:PlenopticMultiplexing:2012} as follows:
\begin{equation}
i_{LF}(x,y) = \int_{u} \int_{v} m(u,v,x,y) \cdot l(u,v,x,y) du dv
\end{equation}
where $i_{LF}(x,y)$ is the intensity measured by the detector and $m(u,v,x,y)$ is the modulation/multiplexing function that encodes the incoming light rays. The modulation function represents the different optical elements that could be used to sense the light field including pinholes ($m(u,v,x,y) = \delta(u,v,x,y)$), Fourier masks, random codes/masks, or diffraction gratings where the modulation functions are Gabor wavelets~\cite{hirsch2014switchable}. Discretizing the above equation, $\textbf{i}_{LF} = \textbf{M}\textbf{l}$ where $\textbf{i}_{LF}, \textbf{l}$ are the vectorized images and light fields, and $\textbf{M}$ is the modulation matrix, and both linear and nonlinear inversions can recover back the light field~\cite{Wetzstein:PlenopticMultiplexing:2012}.
\subsection{Time-of-Flight Imaging}
In contrast, TOF is typically modeled using a cross-correlation between the incoming light signal and the reference code sent to the sensor. Given that incoming light is of the form: $1 + \alpha \cos( f_M t + \phi(x,y) )$ where $\phi$ is the phase accumulated due to the optical path traveled from light source to object to camera and $\alpha$ is the albedo, the intensity at the sensor (normalized to integration time) is:
\begin{equation*}
i_{TOF}(\tau,x,y) = (1 + \alpha(x,y) \cos(f_M t + \phi(x,y) ) ) \otimes cos(f_M t) 
\end{equation*}
\begin{equation}
\approx \frac{\alpha(x,y)}{2} \cos(f_M \tau + \phi(x,y)).
\end{equation}

Here, $\tau$ is the cross-correlation parameter which controls the phase shift of the reference signal. By choosing different $\tau$ such that $f_M \tau = 0, \pi/2, \pi, 3\pi/2$, we can recover both the albedo $\alpha$ and the phase $\phi$ at each spatial location (x,y) using quadrature inversion: 
\begin{equation*}
\phi(x,y) = \tan^{-1}(( i_{TOF}(\frac{3\pi}{2}) - i_{TOF}(\frac{\pi}{2}))/(i_{TOF}(\pi) - i_{TOF}(0) ) ),
\end{equation*}
\begin{equation}
 \alpha =  \sqrt{ (i_{TOF}(\frac{3\pi}{2})-i_{TOF}(\frac{\pi}{2}))^2 + (i_{TOF}(\pi)-i_{TOF}(0))^2 }.
\end{equation}

Note that $d = \frac{c\cdot \phi}{4\pi f_M}$ can directly recover depth $d$ from phase $\phi$ for TOF imaging. 

\subsection{Depth Fields}
We now introduce the concept of the \textbf{depth field} as the ordered pair of albedo and depth (encoded in phase) $(\alpha, \phi)$ that occurs at every $(u,v,x,y)$ spatio-angular coordinate, i.e. $\alpha = \alpha(u,v,x,y), \phi = \phi(u,v,x,y)$. Note that depth fields are not recoverable from TOF measurements alone since TOF assumes a pinhole camera model, which sample $\phi$ and $\alpha$ at a particular fixed $(u,v)$. We now describe the forward model of depth field imaging as follows:

\begin{multline}
i(\tau, x,y) = 
\int_u \int_v m(u,v,x,y)\cdot  \\ (1 + \alpha(u,v,x,y) \cos(f_M t + \phi(u,v,x,y))) du dv \\
 \otimes \cos(f_M t)
\end{multline}
which is approximately
\begin{multline}
i(\tau,x,y) \approx \int_u \int_v m(u,v,x,y)  \cdot\\
\frac{\alpha(u,v,x,y)}{2} \cdot \cos(f_M \tau + \phi(u,v,x,y)) du dv.
\end{multline}

To invert this model, we take four measurements $f_M \tau = 0, \frac{\pi}{2}, \pi, \frac{3\pi}{2}$ to get images $i(0), i(90), i(180), i(270)$ at each spatial location. Then we calculate $\textbf{M}^{-1}i(\tau)$ to invert the light field matrix for each of these images (Note: this inverse can be either done at lower spatial resolution or using sparse priors or modeling assumptions to retain resolution). Thus we recover albedo and phase mixed together at every $(u,v,x,y)$:
\begin{equation}
D' = \frac{\alpha(u,v,x,y)}{2} \cdot \cos(f_M \tau + \phi(u,v,x,y)).
\end{equation}

To unmix the albedo and phase, we can perform quadrature inversion on $D'$ for $f_M \tau = 0, \frac{\pi}{2}, \pi, \frac{3\pi}{2}$ as before in TOF to recover the depth field. 

\section{Methods to Capture Depth Fields}

We describe the potential for single-shot capture of depth fields (Note: single-shot is a misnomer since 4 phase measurements are performed per shot, however such functionality can be built into hardware to work in a single exposure). As in most light field sensors, we can align microlenses above CMOS TOF sensors such as photogates, photonic mixer devices, etc. Doing so allows sampling the angular plane by sacrificing spatial resolution at the sensor plane. The main lens can widen its aperture, allowing more light transmission while each of the sub-aperture views underneath the microlenses maintains a large depth of field~\cite{ng2005light}. This is advantageous since existing TOF cameras sacrifice exposure time to keep a small aperture and large depth of field. One limitation is the need for fine optical alignment of the microlenses at the conjugate image plane in the camera body. 

Another depth field sensor can use amplitude masks between the main lens and the sensor plane of photogates to filter incoming angular rays~\cite{veeraraghavan2007dappled}. While allowing less light transmission as microlenses, masks can be designed with different coding patterns for improved reconstruction of the depth field and can be flexibly interchanged within the camera body unlike fixed optical elements. We note a similar technique from~\cite{godbaz2011extending} which uses a coded aperture in front of LIDAR system to extend the system's depth of field.

We also propose a fully integrated CMOS pixel design that does not require alignment of external optical elements: integrated diffraction gratings over interleaved photogates similar to~\cite{sivaramakrishnan2011enhanced}. This sensor works by diffracting the incoming light to form a Talbot pattern that is imaged by the photogates underneath similar to other diffractive sensors~\cite{hirsch2014switchable}. Note that this pixel can achieve better light efficiency with phase gratings and reduce its pixel size with interleaved photogates while maintaining the advantages of CMOS integration for cost and mass-production. 

\begin{figure}
\begin{center}
   \includegraphics[width=1\linewidth]{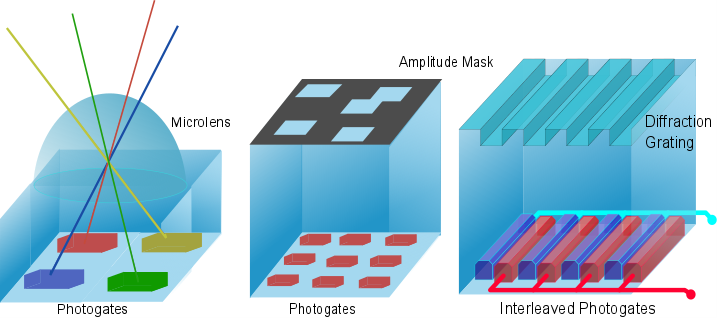}
\end{center}
\caption{Pixel designs for single-shot camera systems for capturing depth fields. Microlenses, amplitude masks, or diffraction gratings are placed over top of photogates to capture light field and TOF information simultaneously.}
\label{pixels}
\end{figure}

We outline the design of these concept pixels in Figure~\ref{pixels}. The only image sensor fabricated to date capable of capturing depth fields in a single shot (that we know of) integrates metal diffraction gratings over single photon avalanche diodes~\cite{Lee2014spad}, however this sensor is used for lensless fluorescence imaging. All these single-shot methods sacrifice spatial resolution to multiplex the incoming depth field. 

Since fabricating a CMOS sensor takes significant time and resources, in this paper we motivate the need for depth field imaging using a custom acquisition setup. This acquisition setup captures depth fields at high spatial resolution by moving a TOF camera on a two axis stage sequentially in the $(u,v)$ plane, as if having an array of TOF cameras, to scan a depth field. See Figure~\ref{conceptfigure} for a schematic depiction of depth field capture.

\section{Experimental Setup}
In this section, we describe the depth field acquistion setup that scans a TOF sensor. We choose this approach since existing TOF sensors have limited resolution so single-shot methods would result in even smaller spatial resolution and the need of precise alignment of microlenses or masks. Our system has some limitations including a bulky setup and static scene acquisition, but we still demonstrate advantages of depth field imaging.  

\begin{figure}
\begin{center}
   \includegraphics[width=1\linewidth]{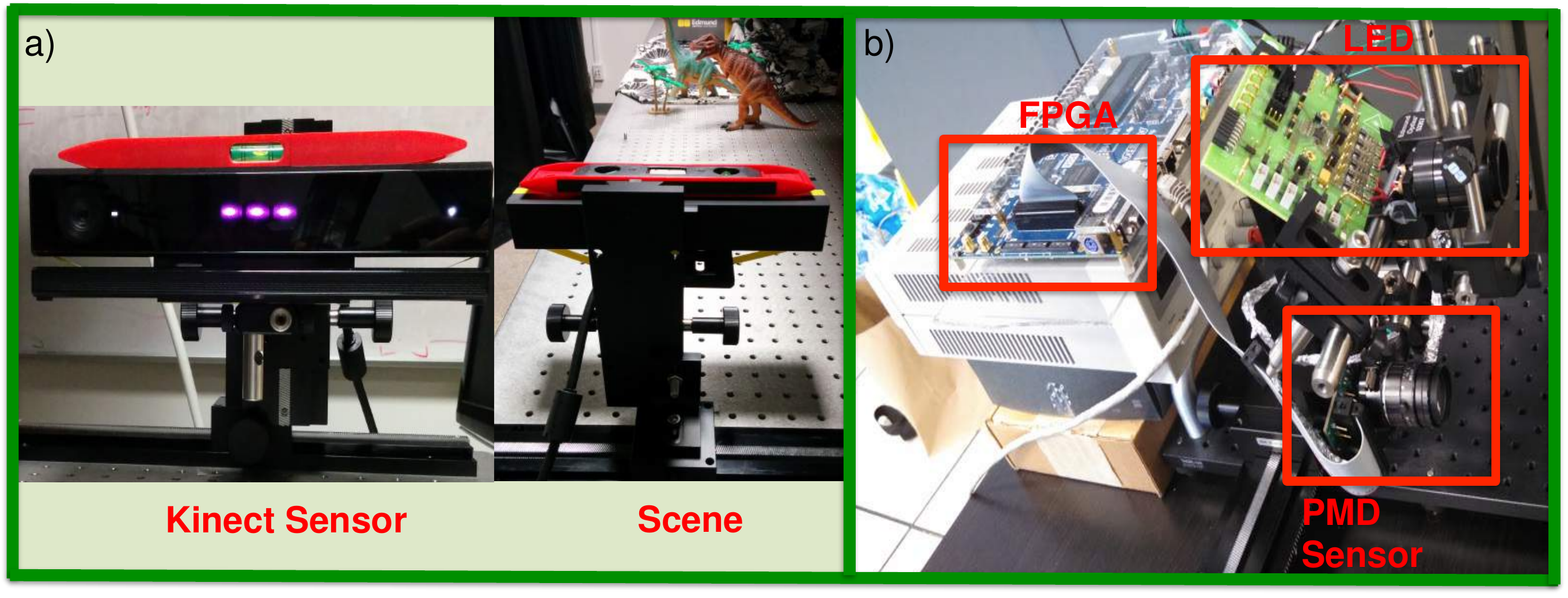}
\end{center}
\caption{Setup to capture depth fields in practice. (a) A Kinect is placed on a XY translation stage on an optical bench, and a representative imaging scene, (b) PMD sensor with FPGA for code generation and LED setup as in ~\cite{kadambi2013coded} }
\label{setup}
\end{figure}

We move a TOF sensor on a two axis stage at different $(u,v)$ positions. We utilize both the Microsoft Kinect One which has a 424 x 512 depth resolution (see Figure~\ref{setup}), and a custom PMD sensor of 160 x 120 resolution which enables us to send custom modulation codes directly to the silicon. The PMD sensor setup is the same as that described in ~\cite{kadambi2013coded}. All depth fields were captured with 1" spacing in the $(u,v)$ plane at the coarse resolution of 5x5. 

\section{Applications of Depth Fields}
\begin{figure*}
\begin{center}
\includegraphics[width=1\linewidth]{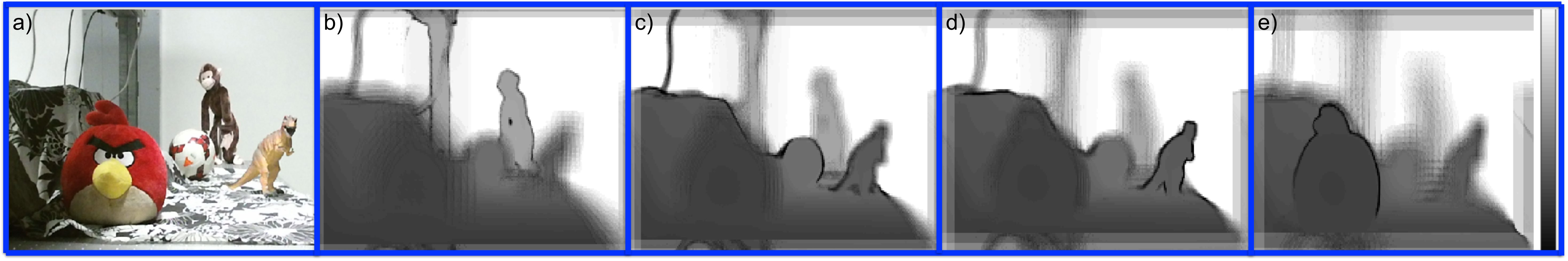} 
\end{center}
   \caption{a) Captured scene, b-e) Digital refocusing on different focal planes for the depth map of the scene, showing how depth field imaging can break the tradeoff between aperture and depth of field for range imaging}
\label{digitalrefocus}
\end{figure*}
In this section, we highlight new applications of depth field imaging. \vspace{-0.5cm}
\subsection{Synthetic Aperture Refocusing}
One main disadvantage of TOF imaging is the necessity of a small aperture for large depth of field to yield accurate depth values. Having a shallow depth of field or wide aperture causes optical blur which corrupts TOF depth values. However, a small aperture limits the shutter speed and increases the acquisition time for these systems. In contrast, light field imaging breaks this tradeoff between depth of field and aperture size by using synthetic aperture refocusing. A plenoptic sensor with microlenses above its pixels can open its aperture and allow more light transmission while keeping the sub-aperture images beneath the microlenses in-focus, albeit at the loss of spatial resolution. After capture, one can digitally refocus the image, thus extending the depth of field by shearing the 4D light field and then summing over $(u,v)$ to synthesize images with different focal planes~\cite{ng2005light}. 

Similarly, we show that the same techniques can be applied to depth fields. In Figure \ref{digitalrefocus}, we show digital refocusing of the 4D $\phi(u,v,x,y)$ information by applying the same shear and then average operation~\cite{ng2005light}. We are able to synthesize capture through a large virtual aperture for the scene which has not been shown in depth maps before, and may be combined with wide aperture light intensity images for enhanced artistic/photographic effect. In addition, this validates that single-shot depth field sensors such as TOF sensor with microlenses can allow more light through the aperture, thus increasing exposure while maintaining the same depth of field. This enables decreased acquisition time for TOF sensors at the expense of computationally recovering the lost spatial resolution and depth of field in post-processing algorithms. We note that~\cite{godbaz2011extending} also showed extended depth of field for a LIDAR system using a coded aperture, but they don't extend their framework to show applications such as digital refocusing.

\subsection{Phase wrapping ambiguities}

One main limitation for single frequency TOF is that the phase has $2\pi$ periodicity, and thus depth estimates will wrap around the modulation wavelength. For modulation frequencies in the tens of MHz, this corresponds to a depth range of a few meters, which can be extended further by using multiple frequencies~\cite{droeschel2010multi, payne2009multiple} or phase unwrapping algorithms~\cite{droeschel2010probabilistic}. However, as modulation frequencies scale higher, phase wrapping becomes more severe. 

We observe that capturing depth fields at a single modulation frequency also allows us to unwrap the phase periodicity by utilizing inherent epipolar geometry from different viewpoints. We use the depth from correspondence algorithm from~\cite{tao2013depth} which is coarse and distance dependent, but does not suffer from phase wrapping, and thus can unwrap the depth measurements given by TOF. 

\begin{figure*}[t]
\begin{center}
\includegraphics[width=0.6\linewidth]{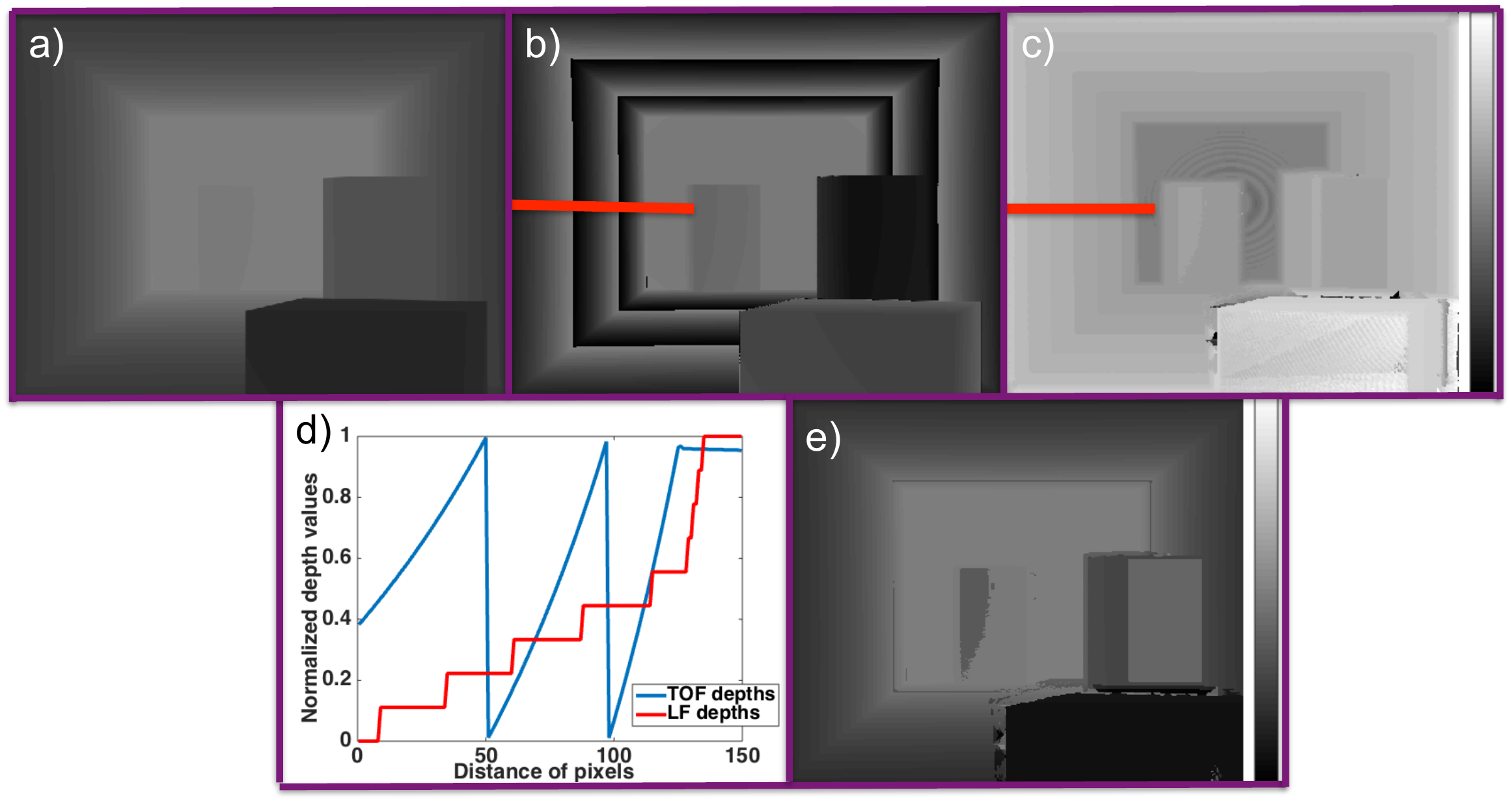} 
\end{center}
   \caption{Phase unwrapping algorithm on synthetic data. a) Cornell box scene with ground truth depth values, b) a phase wrapped scene with red fiducial line for calibration marked, c) depth map given by light field correspondence algorithm. We identify the same calibration line in this scene for phase unwrapping, d) we map the TOF wrapped values to the depth values from correspondence for the given calibration line, e) unwrapped depth map.}
\label{syntheticpw}
\end{figure*}

In Figure~\ref{syntheticpw}, we simulate the Cornell Box scene and capture a depth field using the ray tracer Mitsuba~\cite{jakob2010mitsuba}. We simulate phase wrapping and calculate depth from correspondence. In order to perform phase unwrapping, we select a continuous line in the image (the side wall in this scene) to determine the number of times the TOF image wraps upon itself in the scene. We use this mapping to match the wrapped TOF depth values to the depth values from correspondence, leading to unwrapped TOF depth values for the entire image as shown in Figure~\ref{syntheticpw}d. We also use a median filter to alleviate edge discontinuities in calculating depth from correspondence.

\begin{figure}
\begin{center}
  \includegraphics[width=1\linewidth]{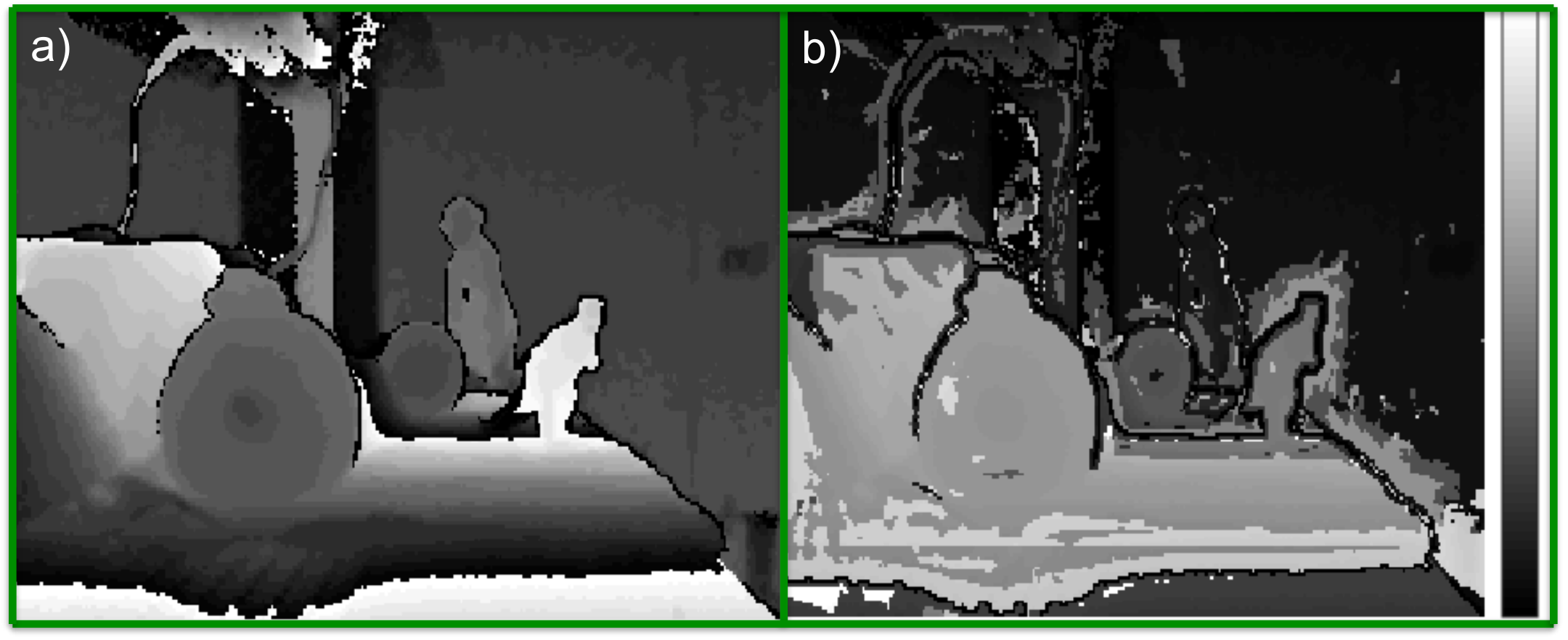}
\end{center}
\caption{a) Phase unwrapping on real data with synthetic phase wrapping induced (due to prototype limitations). b) Recovered depth map. Notice that the monkey in back is not recovered because there does not exist a calibration marker line in the scene that extends all the way back in the TOF image.}
\label{realpw}
\end{figure}

In the Microsoft Kinect we use for capturing depth fields, the maximum modulation frequency is 30MHz, which makes showing phase wrapping difficult on a standard optical bench. Thus we change the bit settings on the Kinect TOF sensors from N bits to N-1 bits to simulate phase wrapping for a real scene (identical to the wrapping caused by periodicity at a higher modulation frequency of 60MHz). We show the results of our phase unwrapping algorithm in Figure~\ref{realpw}. Note that the reconstruction quality is limited by the lack of a good fiducial line in the scene that clearly corresponds light field depths to TOF wrapped depths. This is a limitation of our method, and it would be interesting to explore automatic calibration for phase unwrapping.

\subsection{Refocusing through partial occluders}

\begin{figure*}
\begin{center}
\includegraphics[width=0.6\linewidth]{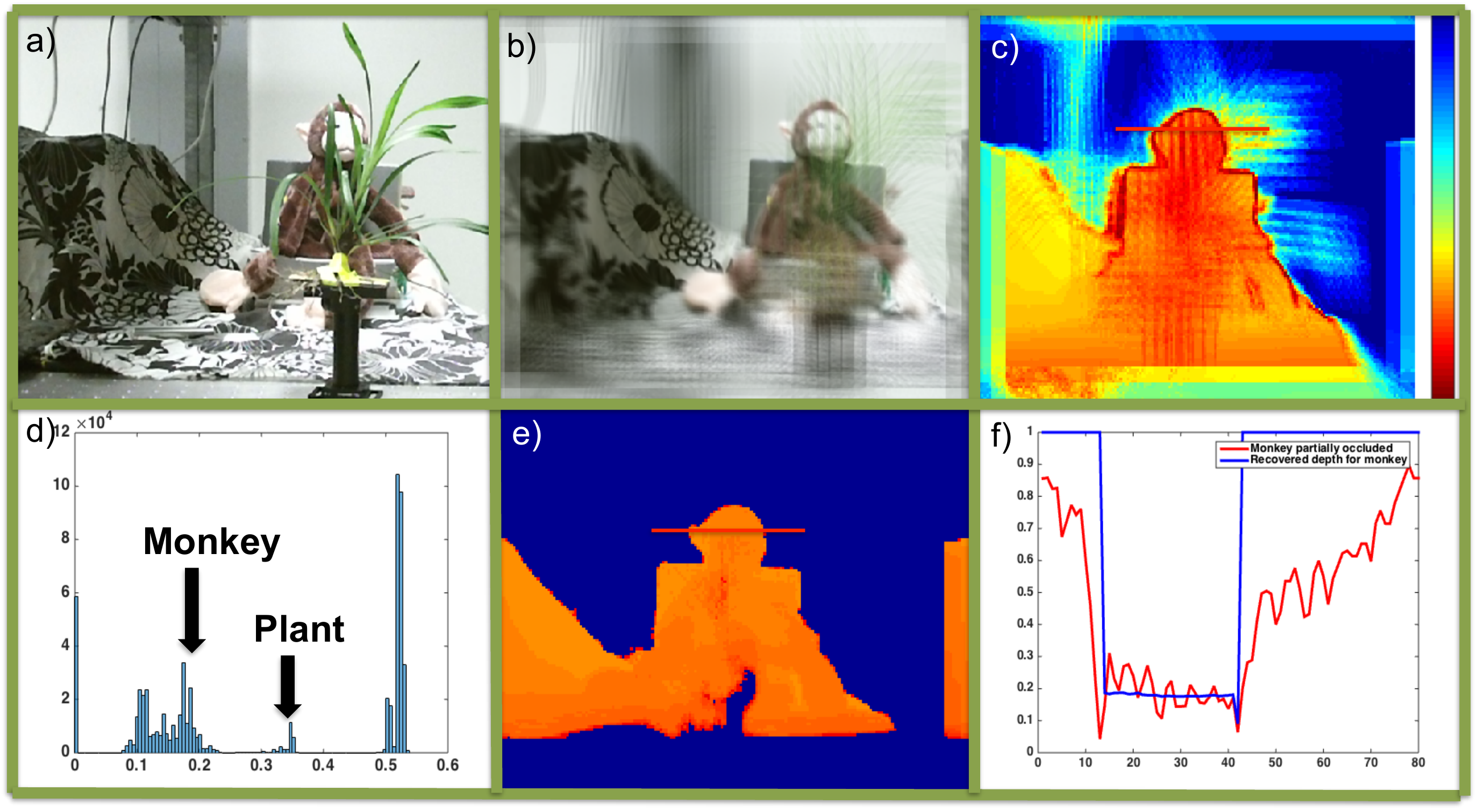} 
\end{center}
   \caption{Refocusing in spite of foreground occlusions: (a) Scene containing a monkey toy being partially occluded by a plant in the foreground, (b) traditional synthetic aperture refocusing on light field is partially effective in removing the effect of foreground plants, (c) synthetic aperture refocusing of depth displays corruption due to occlusion, (d) histogram of depth clearly shows two clusters corresponding to plant and monkey, (e) virtual aperture refocusing after removal of plant pixels shows sharp depth image of monkey, (f) Quantitative comparison of indicated scan line of the monkey's head for (c) and (e)}
\label{partialoccluder}
\end{figure*}

The large synthetic aperture that can be synthesized by capturing 4D depth fields allows us to image past partial occluders in the foreground. This technique, which blurs out the foreground to reveal the background, has been shown in light fields~\cite{vaish2006reconstructing} to look through bushes and plants. Note that in Figure \ref{partialoccluder}, applying the same technique to the depth field works correctly for the albedo (one can see the object clearly while blurring out the foreground), but it does not work for the phase. This is because while visually we can perceptually tolerate some mixing of foreground and background color, this same mixing corrupts our phase measurements, leading to inaccurate depth values. 

To solve this mixing problem when refocusing light fields, researchers have simply not added rays that are from the foreground when averaging over the sheared light field. A key assumption to their algorithm is that the foreground object rays are identified either by shooting continuous video~\cite{wilburn2005high} or by constructing an epipolar image, finding the corresponding depths, and then separating foreground relative to the background~\cite{vaish2006reconstructing}. These algorithms are computationally expensive to identify the occluding objects pixels. We note that~\cite{yang2015} use a combination of unstructured multiview stereo views and a depth sensor to refocus an intensity image through a partial occluder, and use the depth information to create a probabilistic model for occluders. 

In contrast, we utilize the depths directly captured via TOF measurements to construct a histogram of depths observed in the scene as shown in Figure~\ref{partialoccluder}. We then can simply pick a foreground cluster using K-means or another computationally efficient clustering algorithm, which is faster than constructing an epipolar image, estimating line slopes, and then forming a histogram to do clustering. In Figure~\ref{partialoccluder}, you can see the results of our algorithm.

\subsection{Refocusing past scattering media}

\begin{figure*}
\begin{center}
\includegraphics[width=0.7\linewidth]{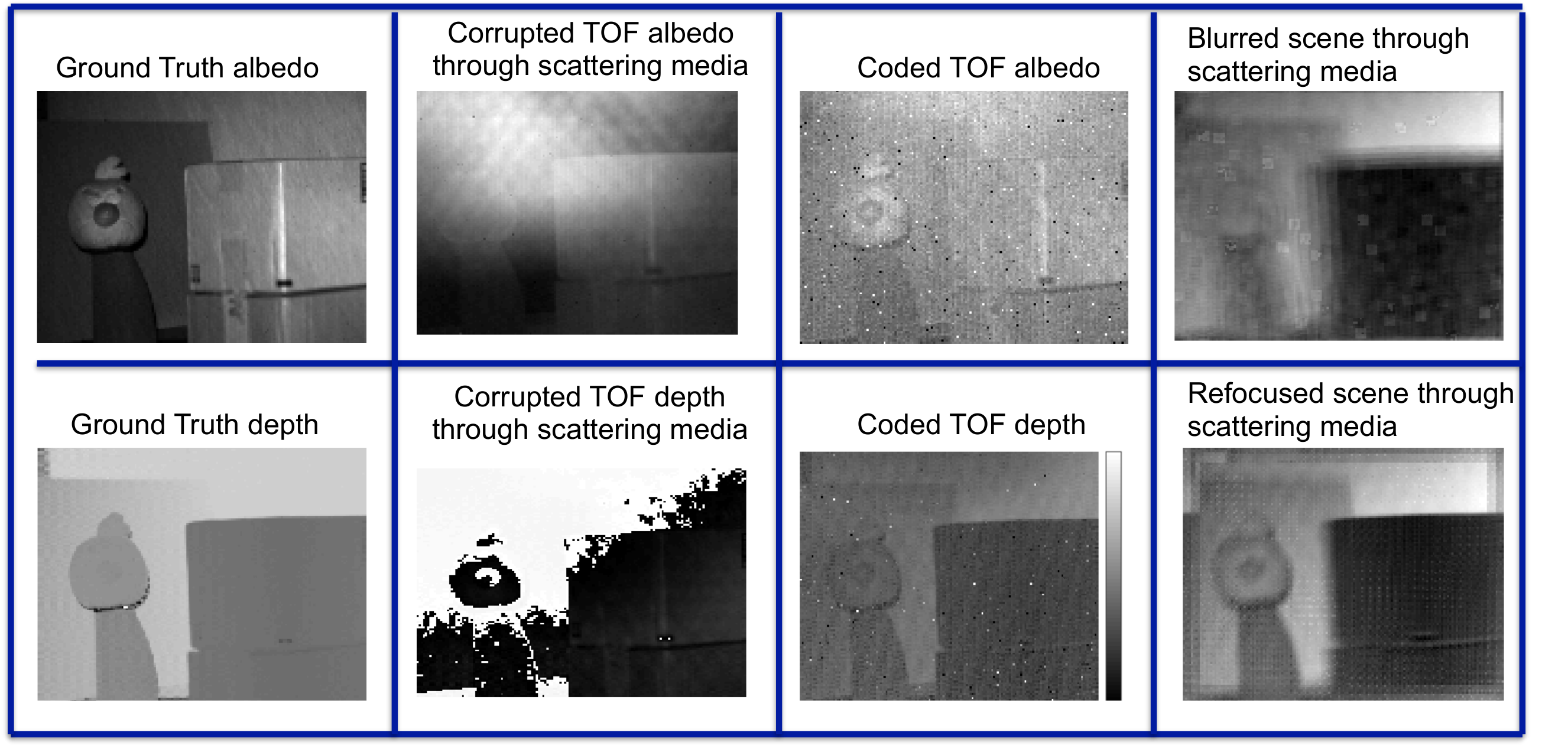} 
\end{center}
   \caption{We use coding techniques from~\cite{Tadano2015} to image beyond backscattering nets. Notice how the corrupted depth maps are improved using the codes. We show how digital refocusing can be performed on the images without the scattering occluders by combining depth fields with coded TOF.}
\label{scattering}
\end{figure*}

While the previous subsection dealt with partial occluders that block the background for certain $(u,v)$ viewpoints, other occluders such as scattering media or translucent objects are more difficult because they mix multiple phase measurements corresponding to different optical path lengths together at a single pixel. We approach the problem via coded TOF, specifically the depth selective codes by \cite{Tadano2015}. Mainly, we show how coded TOF extends the capabilities of our depth field camera systems by imaging past scattering media, and then use spatial information to perform digital refocusing. In Figure~\ref{scattering}, we image through backscattering nets to get a depth field past the scattering media. We place nets in front of the camera to act as strong backscatterers, notice how the depth values are corrupted by the scattering. Using the depth selective codes, we can image past the nets, and using multiple shots at different $(u,v)$ viewpoints, we can capture the depth field beyond the nets and do digital refocusing. This demonstrates how depth field imaging can leverage the advantages of coded TOF techniques, and poses interesting questions of how to design the best possible codes for single-shot depth field imaging systems.


\section{Discussion}

Depth fields unify light field and TOF imaging as a single function of spatio-angular coordinates, and are useful for various applications. Besides the simple extensions of adding two imaging modalities, they can inform each other and make algorithms computationally more efficient and conceptually simpler, particularly in solving the problem of various occluders for light fields by using TOF information and breaking tradeoffs between aperture and depth of field for TOF cameras by adding light field capability. Improvements in light field depth estimation such as in ~\cite{tao2013depth} can also be applied for depth field cameras leading to improved depth resolution.

A key question that concerns depth field cameras is their pixel size which makes pixel multiplexing, including the sensor designs outlined in this paper, problematic. We note that TOF pixels have shrunk currently to 10um~\cite{Payne2014} which is only 10x larger than regular pixels (1um), and that technological advances such as stacked image sensors may help alleviate these multiplexing worries. However, the clear advantages for depth field cameras are applications where spatial resolution is not the limiting factor. This includes imaging systems that are limited by aperture (as argued in Section 6.1) and lensless imaging where spatial pixel layout is not a factor~\cite{gill2011microscale}.

\subsection{Limitations}
Some limitations include long computational algorithms to recover lost spatial resolution for single-shot depth field cameras, or increased acquisition time for large TOF camera arrays or TOF cameras on mechanical gantries to scanline a depth field. Many applications provide partial robustness to depth sensing in the wild, but rely on modeling assumptions (foreground vs. background separation, scattering media is not immersing the object) that limits their deployment in real autonomous systems.

\subsection{Future directions}
One main future direction is to fabricate CMOS photogates with integrated diffraction gratings for an on-chip depth field sensor. This chip would be a lensless depth sensor for biomedical or 3D printing applications where a physical lens is bulky and prevents deployment of the image sensor in confined locations. Our analysis of a depth field shows that one can invert a farfield image using angular resolution similar to lensless lightfield cameras~\cite{gill2011microscale}. 




\section{Conclusion}

We presented the depth field, a representation of light albedo and phase from optical path length as function of space and angle. We developed the forward model to inform new ways to capture depth fields, and showed a myriad of applications that having this information possesess. We are inspired by the possibilitiy of image sensors that can perform hybrid depth imaging in general. 
\newline
\newline
\textbf{Acknowledgements:} The authors gratefully acknowledge Achuta Kadambi for insights into modeling depth field capture and the nanophotography setup, and Ryuichi Tadano for coded TOF experiments and improving phase unwrapping algorithms. This work was partially  supported by NSF CAREER-1150329 and NSF grant CCF-1527501, and S.J. was supported by a NSF Graduate Research Fellowship.

{\small
\bibliographystyle{ieee}
\bibliography{egbib}
}

\end{document}